\documentclass[sigconf,nonacm]{acmart}
\usepackage{tablefootnote}
\usepackage{cleveref}
\usepackage[ruled, linesnumbered, commentsnumbered, longend]{algorithm2e}
\usepackage{tikz}
\usepackage{graphicx}
\usepackage{subcaption}
\usepackage{xcolor}
\usepackage{enumitem}

\begin{document}

\title{Temporal Subspace Clustering for Molecular Dynamics Data}

\author{Anna Beer}
\authornote{Both authors contributed equally to this research.}
\affiliation{%
  \institution{University of Vienna}
  \city{Vienna}
  \country{Austria}}
\email{anna.beer@univie.ac.at}

\author{Martin Heinrigs}
\authornotemark[1]
\affiliation{%
  \institution{LMU Munich}
  \city{Munich}
  \country{Germany}
}
\email{heinrigs-martin@t-online.de}

\author{Claudia Plant}
\affiliation{%
  \institution{University of Vienna}
  \city{Vienna}
  \country{Austria}}
\email{plant@univie.ac.at}

\author{Ira Assent}
\affiliation{%
  \institution{Aarhus University}
  \city{Aarhus}
  \country{Denmark}}
\email{ira@cs.au.dk}

\begin{abstract}
We introduce MOSCITO (MOlecular Dynamics Subspace Clustering with Temporal Observance), a subspace clustering for molecular dynamics data. 
MOSCITO groups those timesteps of a molecular dynamics trajectory together into clusters in which the molecule has similar conformations.
In contrast to state-of-the-art methods, MOSCITO takes advantage of sequential relationships found in time series data. Unlike existing work, MOSCITO does not need a two-step procedure with tedious post-processing, but directly models essential properties of the data. 
Interpreting clusters as Markov states allows us to evaluate the clustering performance based on the resulting Markov state models. 
In experiments on 60 trajectories and 4 different proteins, we show that the performance of MOSCITO achieves state-of-the-art performance in a novel single-step method.
Moreover, by modeling temporal aspects, MOSCITO obtains better segmentation of trajectories, especially for small numbers of clusters.
\end{abstract}


\maketitle

\section{Introduction}
Proteins play an important role in every living organism.
Analyzing the transition between different shapes of a protein, also known as folding, is important for understanding the structure and function of proteins, e.g., for drug design. 
Many diseases like Alzheimer, Parkinson, and different types of cancer are connected to misfolding of certain proteins \cite{hartl2017protein}. Thus, understanding such folding processes may help to prevent or cure diseases. 

Molecular dynamics data contains trajectories of proteins that change their shape over time. It is usually very high-dimensional, with thousands of time steps and hundreds of atoms for a single protein. 
Analyzing molecular dynamics data with traditional clustering is typically not meaningful due to the curse of dimensionality. To mitigate issues with high dimensionality, subspace clustering methods detect clusters in lower-dimensional subspaces~\cite{kriegel2012subspace}.
However, existing subspace clustering methods do not cater for the characteristics of molecular dynamics data.

In this paper, we introduce MOSCITO (\textbf{Mo}lecular Dynamics \textbf{S}ubspace \textbf{C}lustering w\textbf{i}th \textbf{T}emporal \textbf{O}bservance), a subspace clustering algorithm that uses temporal regularization to exploit sequential relationships found in molecular dynamics trajectories that describe folding activities of single molecules. MOSCITO is the first one-step approach to handle this complex data type by identifying the best subspace projection. It finds features tailored specifically to molecular dynamics data. Leveraging methods from video analytics, we show that MOSCITO finds temporally contiguous groups of of protein conformations.  

In trajectories, data points that are near in time are usually more closely related than those that are further apart. 
MOSCITO allows users to choose different sizes of time windows and different weighting methods for temporal regularization: binary, Gaussian, exponential, or logarithmic. 
Interestingly, the resulting clusters can be interpreted as states in a Markov State Model (MSM), which we use for comparing MOSCITO to approaches like SSC, PCA + k-Means, and TICA + k-Means.
As we demonstrate in thorough empirical evaluation, MOSCITO successfully clusters molecular dynamics data in a model that directly captures the essential features and temporal relationships.
Our main contributions are as follows:
\begin{itemize}[left=0pt]
    \item We introduce MOSCITO, a subspace clustering method for molecular dynamics data that makes use of temporal relations between time steps that are close
    \item MOSCITO finds temporally contiguous groups of protein conformations that are similar 
    \item In contrast to currently used methods that require several steps, the clusters found by MOSCITO can \textit{directly} be used as states of a Markov State Model describing the trajectory, yielding state-of-the-art quality with a one-in-all holistic approach
\end{itemize}
\section{Clustering trajectories of molecular dynamics data} \label{sec:cluster_md}

In this paper, we study the trajectories of a single molecule over time. A trajectory is given by the three-dimensional Cartesian coordinates of all atoms of the molecule for a number of time steps with fixed intervals.
Where most traditional clustering methods target numerical data in $\mathbb{R}^d$, clustering MD data brings some unique challenges as we outline in the following. 

Due to the size of MD data, automated methods for analyzing them are required. These often involve clustering methods to automatically group similar groups of atoms or timesteps.

In this paper, we focus on simulations of very long molecular dynamics trajectories that show folding processes of single molecules~\cite{shaw_md}. 
Based on the starting positions of a molecule's atoms, physical forces on all other atoms can be calculated, yielding simulated trajectories with a temporal resolution that usually lies in the femtosecond range.
While the overall shape of the molecules is decisive for its function, several features are shown to be relevant, e.g., the torsions between $C^\alpha$ atoms that form the backbone of the protein, distances between specific atoms, or the solvent accessible surface area.

There are different use cases for which a clustering of MD trajectories is required. 
E.g., clusters of atoms that move similarly over time can be used to find dynamic domains, i.e., fragments of a protein that are internally rigid \cite{romanowska2012determining}. 
Clustering time steps in which the conformation of the protein, i.e., the arrangement of its atoms in space, is similar is used for detecting relevant states of the protein. 
Those states can be linked to (local) valleys in the energy landscape of the protein and are e.g. used for developing Markov models that describe a protein trajectory. 
Thus, one goal is to separate metastable states of a protein and the transitions between those states.

\section{Markov state modeling} \label{sec:msm}

Markov state models (MSMs) can be used to describe the dynamics of a MD trajectory by using a $n \times n$ transition matrix. To generate an MSM from an MD trajectory $X=\{x_1,...,x_t\}$, 
discrete states of the trajectory need to be identified. 
State-of-the-art methods use a two-step approach in which they first reduce the dimensionality of the data by using PCA or TICA and then clustering it with $k$-Means into $n$ discrete states $S=\{S_1,...S_n\}$\cite{pyemma_tutorial}. These states are used to calculate the transition probability matrix for a given lag time $\tau$ by counting the transitions from time step $t$ to $t+\tau$: 
\begin{equation}
    p_{ij}(\tau)=\mathbb{P}(x_{t+\tau}\in S_j|x_t\in S_i)
    \label{msm_transition_func}
\end{equation}
The resulting MSM can be used for further analysis of the trajectory. In this paper, we analyze the clustering performance of our method in comparison to state-of-the-art methods for clustering MD trajectories based on their suitability as Markov states.

The variational approach for Markov models (VAMP)\cite{vamp} is a generalization of the \textit{Variational approach to conformational dynamics} (VAC) \cite{nuske2014variational}, \cite{noe2013variational} and provides a tool to evaluate MSMs. The VAC allows the approximation of eigenvalues and eigenfunctions with statistical observables \cite{noe2013variational}. It also defines a family of scoring functions, called the $VAMP-r$, which can be calculated from data. This score can be used to compare the choices of hyper-parameters, like the featurization of the data or the used clustering, used for constructing the MSM. The score is calculated from the singular value decomposition of the Koopman operator. The Koopman operator is explained below. It is important to note that the VAMP score can only be compared for MSMs constructed with the same lag time $\tau$ and cannot be used to find the optimal lag time $\tau$ itself.

The Koopman operator gives the conditional expected value of an arbitrary observable $g$ at time $t+\tau$ for a given $x_t$. The Koopman operator $K_\tau$ of a Markov process is defined as follows \cite{vamp}:
\begin{equation}
    \textit{K}_\tau g(x)=\mathbb{E}[g(x_{t+\tau}|x_t=x]
    \label{eq:Koopman_operator}
\end{equation}
When choosing the Dirac delta function $\delta_y$ centered at y, applying the Koopman operator evaluates the transition density of the dynamics \cite{vamp}.
\begin{equation}
    K(\tau)=C_{00}^{-\frac{1}{2}}C_{01}C_{11}^{-\frac{1}{2}}\approx U_m\Sigma_mV_m^T,
    \label{eq:Koopman_approx}
\end{equation}
where $\Sigma_m=diag(\sigma_1,...,\sigma_m)$ are the first $m$ singular values. 

\paragraph*{VAMP-r Score}
The VAMP-score \cite{vamp} defines a function that can be used to score the quality of MSMs for a set lag-time $\tau$.  This score can be used to compare the methods used to create the discrete trajectories from the same featured trajectory \cite{msm_abilities}.

With $\Sigma_m$ as defined above (see \Cref{eq:Koopman_approx}), the VAMP-r score can be written as:
\begin{equation}
    \text{VAMP-}r=\sum_{i=1}^m \sigma_i^r
    \label{eq_VAMP-r}
\end{equation}

The VAMP-2-score is based on the sum of squared eigenvalues of the transition matrix and can be interpreted as kinetic content~\cite{msm_abilities}. In this paper, we use the VAMP-2 score to score the performance of different discretization methods for MD trajectories since there usually is no given ground truth to compare to. The most commonly used scores are the VAMP-1, which is analogous to the Rayleigh trace, and the VAMP-2, which is analogous to the kinetic variance~\cite{feature_sel}.

\section{Subspace clustering} \label{sec:sc}
In high-dimensional datasets where the curse of dimensionality renders traditional clustering infeasible, subspace clustering finds clusters in a union of lower dimensional subspaces \cite{survey_sc}. For a dataset $X=[x_1, x_2, ..., x_n]\in \mathbb{R}^d$ it is often assumed that vectors $x_j$ are drawn from a union of $k$ subspaces $\{S_i\}_{i=1}^k$ of unknown dimensionality \cite{li2015tsc}. As MD data is typically very high dimensional, subspace clustering is a suitable approach.
For details on subspace clustering, see \cite{parsons2004subspace, vidal2011subspace}.
In this paper, we focus on spectral-based subspace clustering.

The advantages of spectral subspace clustering methods include a relatively simple implementation and robustness regarding data corruption and initialization \cite{survey_sc}. Spectral subspace clustering methods usually perform two steps. First, a subspace representation of the original data is learned and used to construct an affinity matrix. Then spectral clustering is used on the affinity matrix to obtain the final clustering. Many different spectral algorithms like sparse subspace clustering (SSC) \cite{SSC} or Low-Rank Representation (LRR) \cite{liu2012lrr} have been proposed, for an overview see \cite{vidal2011subspace}.
While spectral clustering has been successfully applied to numeric data, it lacks the ability to handle temporal data. We, thus, turn our attention to methods that have been developed with a different application goal in mind, but, as we argue, with similarities that make them attractive for the design of our MD subspace clustering method.

Temporal subspace clustering (TSC)~\cite{li2015tsc} is a subspace clustering algorithm designed to segment video sequences into separate human motions. 
Video sequences are high-dimensional time series, with one frame of the video at each timestep. 
Significant changes in a video sequence usually are not immediate, but take place over many frames. 
Neighboring time frames are more correlated to each other than more distant frames. 
In that sense, molecular dynamics data is similar to video data. 
Unlike classical subspace clustering methods, which do not take any relations between the dimensions into account, TSC is designed to take advantage of the sequential relationship in time series data. 
TSC uses a dictionary learning approach together with a temporal Laplacian regularization function to encode those relationships in the dataset. Dictionary learning is a method that learns a set of basis vectors, a so-called dictionary \cite{tovsic2011dictionary}. The original data can then be represented as linear combinations of the basis vectors. TSC combines the dictionary learning approach with a temporal regularization function to encode the sequential relationships in the data. 
A non-negative dictionary and a corresponding coding are learned from the given data creating a robust affinity graph.

\section{Moscito: Temporal Subspace Clustering for MD Data} \label{chap:method} 
\begin{figure}
    \centering
    \includegraphics[width=.9\linewidth]{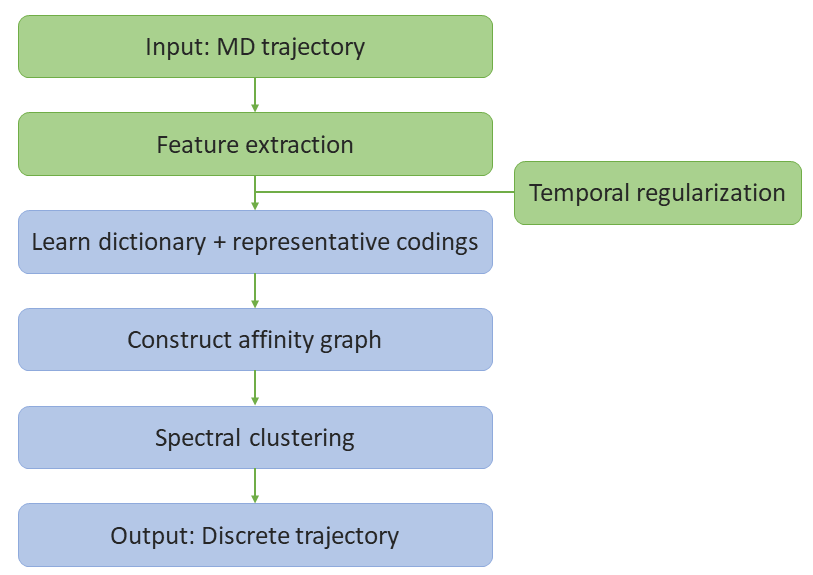}
    \caption{MOSCITO derives MD features from input trajectories. Temporal regularization provides the means for effective subspace clustering with temporal relations.}
    \label{fig:MOSCITO_overview}
\end{figure}
We are now ready to describe our MOSCITO (\textbf{MO}lecular dynamics \textbf{S}ubspace \textbf{C}lustering with \textbf{T}emporal \textbf{O}bservance) approach, the first subspace clustering method for molecular dynamics data that captures temporal relations. 
Existing methods~\cite{pyemma_tutorial} often rely on a two- or three-step approach consisting of 1) dimensionality reduction, 2) clustering, and 3) finding macrostates by combining several clusters. As these steps are handled in isolation, their separate results may not be optimal overall. To avoid such undesirable side effects,  MOSCITO takes a holistic approach: 
Firstly, in subspace clustering, it inherently combines clustering with dimensionality reduction. 
Secondly, MOSCITO incorporates temporal relation by weighting sequential neighbors in a time window. In contrast, existing subspace clustering algorithms cannot capture e.g., correlation between consecutive time steps.

For our new method MOSCITO, we exploit the similarity between MD data and video data (see \Cref{sec:sc}) as well as the advances in the field of data mining on videos: TSC~\cite{li2015tsc} clusters video data while incorporating the temporal relations between time steps as well as allowing clusters in subspaces of the data. Thus, we base MOSCITO on TSC. However, TSC is not able to handle MD data, because despite some similarities between the videos and MD data (e.g., smooth temporal transitions between time steps and potential clusters in subspaces), they also have strong differences: for every time step , MD data describes a chain of atoms with its 3d coordinates, while video data contains colors for every pixel, where pixels are in a 2d grid. 
Thus, we need to change several steps and include knowledge from the field of MD as shown in Figure \ref{fig:MOSCITO_overview}, which provides an overview of MOSCITO. 
The blue parts are equivalent to those of TSC \cite{li2015tsc}, see \Cref{chap:rel_work}. 
The green parts are adapted or changed completely. 
At first, essential MD features 
are extracted from the MD trajectory. 
From the features and a temporal regularization function, we learn a dictionary and the corresponding representation coding matrix. 
We then construct an affinity graph, and generate temporal segments by clustering the affinity graph.

\paragraph*{Feature extraction} 
From the input MD trajectors, we extract features that capture essential properties. \\
\textbf{Cartesian coordinates:} 
The $C^\alpha$ atoms determine the shape and function of a protein, thus we use the common approach to focus solely on their 3d coordinates, rather than considering all atoms. 
For each frame,  coordinates are centered and aligned to the first frame of the trajectory.\\
\textbf{Backbone torsions:} 
The backbone torsions determine the main structure of a protein. The torsion angles $\phi$ and $\psi$ describe the dihedral angles between atoms of the backbone. Backbone torsions better describe the structure of a protein than absolute coordinates \cite{vaidehi2015internal} as internal angles are independent of the global rotation of the protein, removing the need to align all frames to a fixed rotation.  \\
\textbf{Distance based:} Heavy atoms are all atoms that are not hydrogen. We extract the minimal distance between heavy atoms \cite{BONOMI20091961} (we exclude the pairs at indices(i, i+1) and (i, i+2), as their distance is largely determined by their chemical bonds). Following \cite{feature_sel}, distances $d$ are transformed to the negative exponential function $e^{-d}$  \\
\textbf{Flexible torsions:} The flexible torsion feature consists of the $\chi_1$ to $\chi_5$ side chain torsions. The side chains of a protein are chains of atoms, connected to the backbone. The angles are transformed into sine and cosine angles.\\
\textbf{Solvent accessible surface area:} For each frame of the trajectory, the solvent accessible surface area (SASA) of each residue is calculated using the Shrake-Rupley algorithm \cite{SHRAKE1973351}. The SASA of a residue is calculated by summing up the SASA of each of the residues atoms\\
\textbf{3D shape histogram:} Based on \cite{ankerst19993d}, a combined 3D shape histogram for each frame of the trajectory is created. The state space is separated into five equally spaced shells around the origin of the coordinate system. The shells are split up further by combining them with 12 sectors into 60 cells. Histograms are created by counting the number of atoms in each cell. Unlike \cite{ankerst19993d}, all atoms instead of only the surface atoms were used as we are not only interested in the shape and function of the protein at a certain time step, but also in its internal folded structure. Final histograms are normalized between 0 and 1.
\paragraph*{Temporal regularization}
For trajectories, data points at neighboring time steps are usually more closely related than those at more distant time steps. Let the $i$-th column of the coding matrix $Z$ a representation of vector $x_i$, then the encodings of its sequential neighbors, e.g., $z_{i-1}, z_{i+1}$ should be similar to $z_i$. To achieve this, we use a temporal Laplacian regularization function $f(Z)$. 

\begin{equation}
    f(Z)=\frac{1}{2}\sum_{i=1}^n \sum_{j=1}^n w_ij ||z_i - z_j||_2^2 = tr(ZL_T Z^T)
    \label{eq:TSC_temp_reg}
\end{equation}
$L_T$ is the temporal Laplacian matrix defined as $L_T = \tilde D - W$. $W$ is the weight matrix, capturing the sequential relationship in $X$, and $\tilde D_{ii} = \sum_{j=1}^nw_{ij}$ is a diagonal matrix with the sum of the rows of $W$ on its diagonal.

The temporal regularization in MOSCITO weighs the sequential neighbors using a weight matrix $W\in\mathbb{R}^{n\times n}$ for a trajectory with $n$ timesteps. 
Row $i$ of $W$ represents the weights of $x_i$'s neighbors. 
In the base case, binary weights are used. 
This weighs all $s$ sequential neighbors of $x_i$ the same, independent of the actual temporal distance between the data points. 
In contrast to TSC, MOSCITO offers several 
Naturally, points in close temporal proximity are more strongly correlated than points further apart in time. To model the decrease in correlation, we introduce different decaying weightings: \\
\textbf{Binary:}  weights of 1 for points in the neighborhood and 0 outside\\
\textbf{Gaussian:} Gaussian curve fitted to the neighborhood\\
\textbf{Logarithmic:} Logarithmic decay with a value of 1 for the direct neighbors dropping logarithmically to 0 for $x_{i+s}$ and $x_{i-s}$\\
\textbf{Exponential:} Exponential decay w.r.t. the temporal distance 
\paragraph*{Dictionary learning} Similar to TSC~\cite{li2015tsc}, MOSCITO learns to represent dataset $X$ by a dictionary $D$ and a coding matrix $Z$, such that $X\approx DZ$ \cite{li2015tsc}. 

Adopting a least square regression approach to dictionary learning, we use the following objective function~\cite{li2015tsc}
\begin{equation}
    \begin{split}
        \min_{Z,D}||X-DZ||_F^2+\lambda_1||Z||_F^2\\
        \text{s.t. } Z \geq 0, D \geq 0, ||d_i||_2^2 \leq 1,
    \end{split}
    \label{eq:tsc_obejctive_lsr}
\end{equation}
where $||Z||_F^2$ denotes the Frobenius norm, and we use $\lambda_1$ for balancing both terms. 

Importantly, MOSCITO integrates temporal regularization to capture relationships between nearby time steps. We achieve this by adding temporal regularization function $f(Z)$ such that our new objective function is the same as in TSC~\cite{li2015tsc}
\begin{equation}
    \begin{split}
        \min_{Z,D}||X-DZ||_F^2+\lambda_1||Z||_F^2 + \lambda_2f(Z) \\
        \text{s.t. } Z \geq 0, D \geq 0, ||d_i||_2^2 \leq 1,
    \end{split}
    \label{eq:tsc_obejctive}
\end{equation}
with $\lambda_2$ tradeoff parameter~\cite{li2015tsc}.

To solve the dictionary learning problem efficiently, we follow the alternating direction method of multipliers (ADMM) \cite{li2015tsc}. It decomposes the problem into subproblems that are more easy to solve. Using auxiliary variables $U, V$, Eq. \ref{eq:tsc_obejctive} is equivalent to
\begin{equation}
    \begin{split}
        \min_{Z,D,U,V}||X-UV||_F^2+\lambda_1||V||_F^2 + \lambda_2f(V) \\
        \text{s.t. } U=D, V=Z, Z \geq 0, D \geq 0, ||d_i||_2^2 \leq 1.
    \end{split}
    \label{eq:MOSCITO_obejctive}
\end{equation}
The augmented Lagrangian of Eq. \ref{eq:MOSCITO_obejctive} is:
\begin{equation}
    \begin{split}
        \mathcal{L} & = \frac{1}{2}||X-UV||_F^2 + \lambda_1||V||_F^2 + \lambda_2 tr(VL_TV^T) \\
        & + \langle \Lambda, U_D\rangle + \langle \Pi, V-Z\rangle + \frac{\alpha}{2}||U-D||_F^2 + \frac{\beta}{2}||V-Z||_F^2,
    \end{split}
    \label{eq:MOSCITO_lagrangian}
\end{equation}
with $\Lambda$ and $\Pi$ Lagrangian multipliers \cite{li2015tsc}, which can now be solved by alternatingly minimizing $\mathcal{L}$ respective to $V, U, Z$ and $D$. 

\textbf{Updating $\bf V$:} To find the minimum of $\mathcal{L}$, its derivative with respect to $V$ is set to zero, resulting in the following equation \cite{li2015tsc}:
\begin{equation}
    [I\otimes (U^TU+(\lambda_1+\alpha)I) + \lambda_2L_T\otimes I] vec(V)=vec(U^TX-\Pi+\alpha Z)
    \label{eq:TSC_V}
\end{equation}

\textbf{Updating $\bf U$:} As for $V$, set derivative of $\mathcal{L}$ wrt. $U$ to zero \cite{li2015tsc}:
\begin{equation}
    U=(XV^T-\Lambda+\alpha D)(VV^T+\alpha I)^{-1}
    \label{eq:TSC_U}
\end{equation}

\textbf{Updating $\bf Z$ and $\bf D$:} The update rules for $Z$ and $D$ are \cite{li2015tsc}: 
\begin{equation}
    Z=F_+(V+\frac{\Pi}{\beta})
, \qquad
    D=F_+(U+\frac{\Lambda}{\alpha})
    \label{eq:TSC_D}
\end{equation}
where $(F_+(A))_{ij}=max\{A_{ij}, 0\}$, non-negativity constraint. Additionally, each column of $D$ is normalized to unit length  $||d_i||_2^2\leq1$.

\paragraph*{Clustering}
Using the learned coding matrix $Z$, an affinity graph is constructed. To capture sequential relationships, the following similarity measure is used \cite{li2015tsc}: 
\begin{equation}
    G(i, j)=\frac{z_i^T z_j}{||z_i||_2||z_j||_2}
    \label{eq:tsc_G}
\end{equation}
Spectral clustering on graph $G$ yields the final clusters.

\begin{algorithm}
\caption{MOSCITO}
\SetKwData{Left}{left}\SetKwData{This}{this}\SetKwData{Up}{up}
\SetKwInOut{Input}{input}\SetKwInOut{Output}{output}

\Input{\textrm{Trajectory $traj$, dictionary size $d$, sequential neighbors $s$, number of clusters $k$, parameters $\lambda_1$, $\lambda_2$, $\alpha$, $\beta$}}
\Output{Discrete trajectory vector $d\_traj$}
\BlankLine
Extract features $X$ from traj\;
\While{not converged} 
{
    Update $V_{k+1}$ according to Eq. \ref{eq:TSC_V} while fixing others\;
    Update $U_{k+1}$ according to Eq. \ref{eq:TSC_U} while fixing others\;
    Update $Z_{k+1}$ according to Eq. \ref{eq:TSC_D} while fixing others\;
    Update $D_{k+1}$ according to Eq. \ref{eq:TSC_D} while fixing others\;
    Update $\Pi_{k+1}: \Pi_{k+1}=\Pi_{k}+\nu \alpha (V_{k+1}-Z_{k-1})$\;
    Update $\Lambda_{k+1}: \Lambda_{k+1}=\lambda_k+\nu \beta (U_{k+1}-D_{k-1})$\;
    $k=k+1$\;
}
Generate affinity graph $G$ using Eq. \ref{eq:tsc_G}\;
Perform spectral clustering on $G$, get $d\_traj$\;
\end{algorithm}

\section{Experiments} \label{chap:experiments}

\paragraph{Implementation} 
We extract features of MD trajectories using libraries \textit{PyEMMA} \cite{scherer_pyemma_2015} and \textit{MDTraj}\footnote{\url{https://www.mdtraj.org}}.
Matrices are represented as \textit{NUMPY} arrays and sparse matrices by \textit{scipy} sparse matrices. 
We efficiently solve linear equations of the form $Ax=b$ using the \textit{PyPardiso 0.4.2}\footnote{\url{https://github.com/haasad/PyPardisoProject}} (version 2021.4) wrapper for the Intel oneMKL PARDISO library. 
For spectral clustering, we use the \textit{scikit-learn} implementation. 
Our Python code for MOSCITO is \href{https://github.com/Mhae98/MOSCITO}{publicly available} \footnote{\url{https://github.com/Mhae98/MOSCITO}}. 
For PCA, TICA and $k$-Means we use the implementations in \textit{PyEMMA}.
The implementation of SSC is based on \cite{SSC_MD} and a Python translation of SSC \footnote{\url{https://github.com/abhinav4192/sparse-subspace-clustering-python}} 
All experiments run on a virtual machine with 64 CPUs and 512GB of RAM running Ubuntu 22.04.

\paragraph{Datasets and parameter settings}
We use MD trajectories of four proteins: 2F4K, Ace, 2WXC, and Savinase, which differ in the number of atoms $\mid A\mid$, number of C$\alpha$ atoms $\mid C\alpha\mid$, and number of residues $\mid R\mid$, as summarized in Table \ref{tab:trajectories}, together with number of  trajectories and timesteps in each trajectory.
If not specified otherwise, parameter default values are 
 \(d= 60\), \(s= 3\), \(\lambda_1= 0.01\), \(\lambda_2= 15\), \(a= 0.1\), \(b= 0.1\).

\begin{table}[ht]
    \centering
    \begin{tabular}{c|c|c|c|c|c|c}
         Protein & $\mid$A$\mid$ & $\mid$C$\alpha\mid$ & $\mid$R$\mid$ & $\mid$T$\mid$ & \# Trajs & Source \\
         \hline 
         \hline
         2F4K & 577 & 35 & 68 & 10.000 & 20 & \cite{shaw_md}\\
         Ace & 264 & 21 & 23 & 10.000 & 28 & \cite{McGibbon2014} \\
         2WXC & 710 & 47 & 47 & 10.000 & 10 & \tablefootnote{\label{foot:source_traj}Birgit Schiøtt’s Biomodelling group, Department of Chemistry, Aarhus University}\\
         Savinase & 3.725 & 269 & 269 &  $\sim$1500 & 2 & \footref{foot:source_traj}
    \end{tabular}
    \caption{Molecular dynamics data}
    \label{tab:trajectories}
\end{table}

\subsection{MOSCITO settings and features} \label{sec:MOSCTIO_params}
In this section, we analyze the parameter sensitivity and impact of feature extraction for MOSCITO. 
For the following experiments, 10 trajectories of the 2WAV-protein are used. The Markov state models are constructed for $k\in\{10, 50, 100, 200\}$ clusters with a lag time $\tau\in\{1, 10, 100\}$. The average VAMP2-score of the 10 trajectories is calculated for each choice of clusters $k$ and lag time $\tau$. The VAMP2-score is calculated for five eigenvalues.
It is important to keep in mind that the VAMP2-score for different MSM lag times is not directly comparable, only for different numbers of clusters.

\subsubsection{Dictionary size} \label{exp:d_size} 
\begin{figure}[ht]
    \centering
    \includegraphics[width=\linewidth, trim= {20 50 20 20},clip]{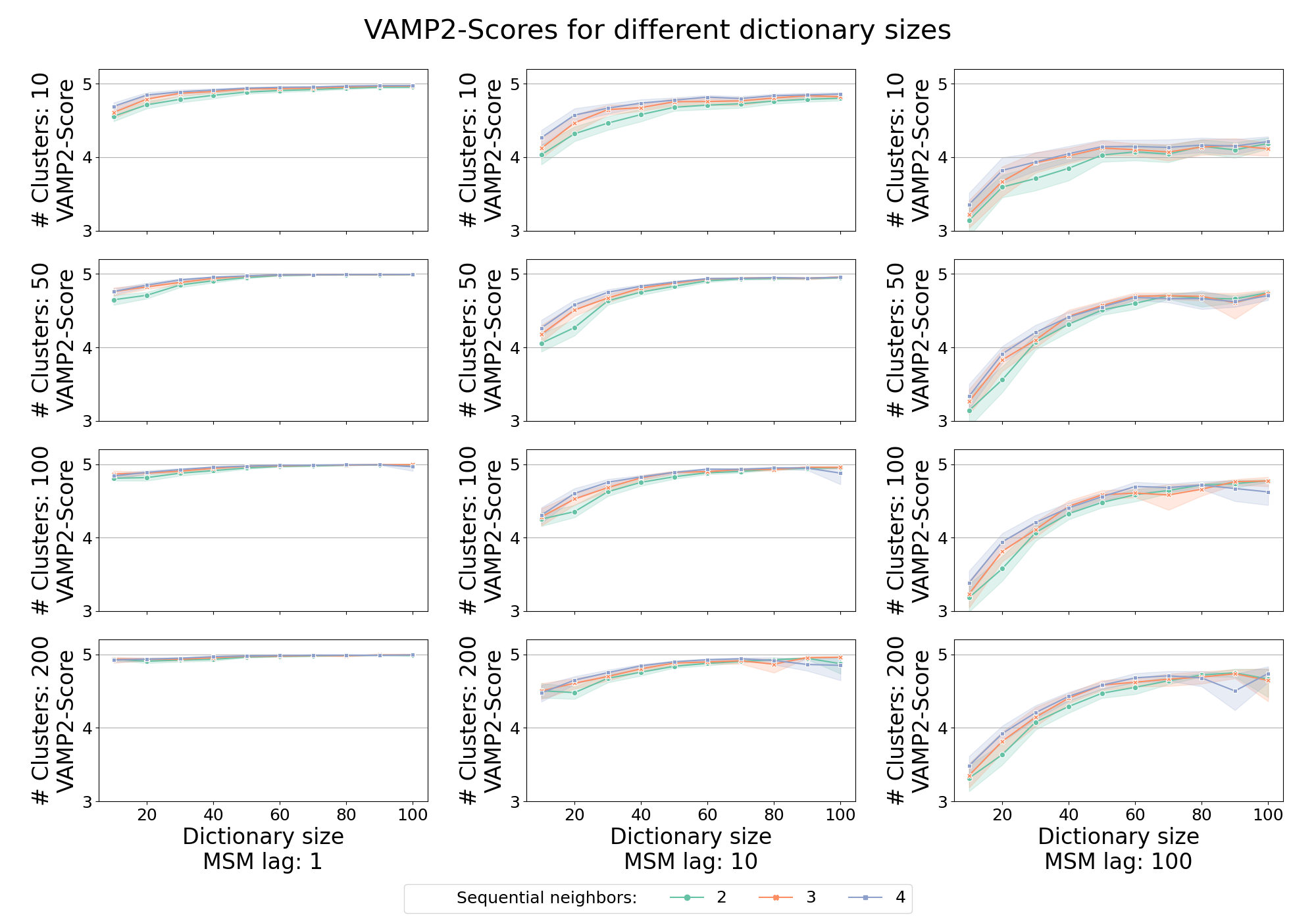}
    \includegraphics[width=\linewidth, trim= {400 5 400 965},clip]{images/experiments/TSC-parameters/plot_d_size.png}
    \caption{Comparison of the VAMP-scores for different dictionary sizes.  For each combination of MSM lag and cluster count, the VAMP2-Score for different dictionary sizes and the number of sequential neighbors are compared. For better differentiation, the VAMP2-score is plotted starting at 3.}
    \label{fig:tsc_param_d_size}
\end{figure}

$d$ defines the dimensionality of the learned dictionary and thus the number of subspaces. As such, it is crucial for the clustering and runtime performance of MOSCITO, where larger dictionary size $d$ provides more subspaces to represent the data.
We study $d\in\{10, 20, ..., 100\}$, and additionally $s\in\{2, 3, 4\}$ number of sequential neighbors. Figure \ref{fig:tsc_param_d_size} shows VAMP2-scores for different numbers of clusters (rows) and MSM lag times (columns): clustering performance improves for larger dictionary size $d$ until it converges. For smaller MSM lag times, the VAMP2-score converges faster than for larger lag times, independent of the number of clusters. With an increasing number of sequential neighbors $s$, the overall shape of the curves is highly similar, with a slight preference for more neighbors that capture more of the temporal relationships.

\subsubsection{Sequential neighbors} \label{exp:seq_n}
\begin{figure}[ht]
    \centering
    \includegraphics[width=\linewidth, trim= {20 50 20 20},clip]{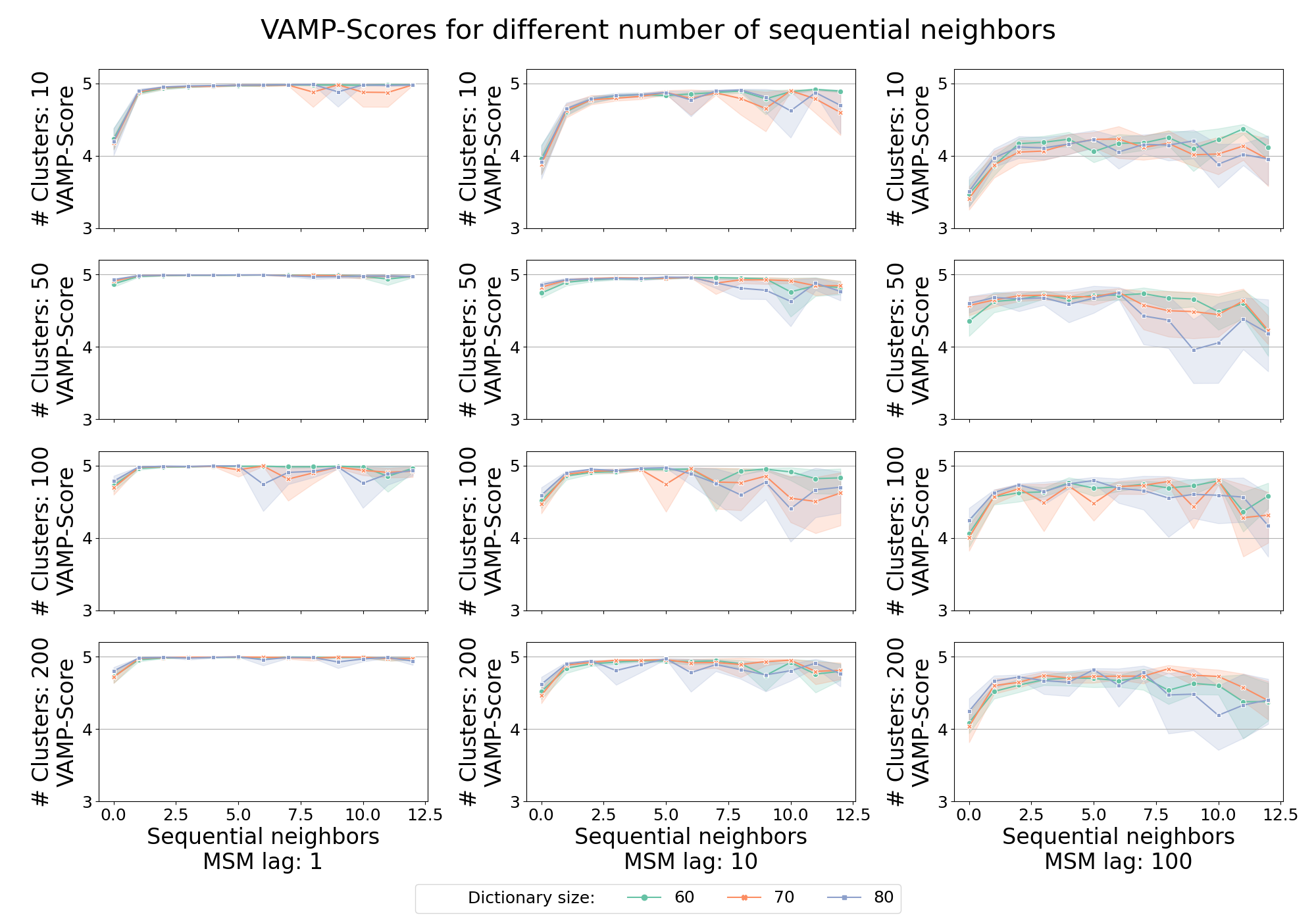}
    \includegraphics[width=\linewidth, trim= {400 5 400 965},clip]{images/experiments/TSC-parameters/plot_seq_neighbors.png}
    \caption{VAMP-scores for varying number of sequential neighbors and dictionary sizes over pairings of MSM lag and cluster count. For visibility, VAMP2-axis starts at 3. Best results for 3 to 5 sequential neighbors.}
    \label{fig:tsc_param_seq_neighbors}
\end{figure}

\begin{figure}[ht]
    \centering
    \includegraphics[width=\linewidth]{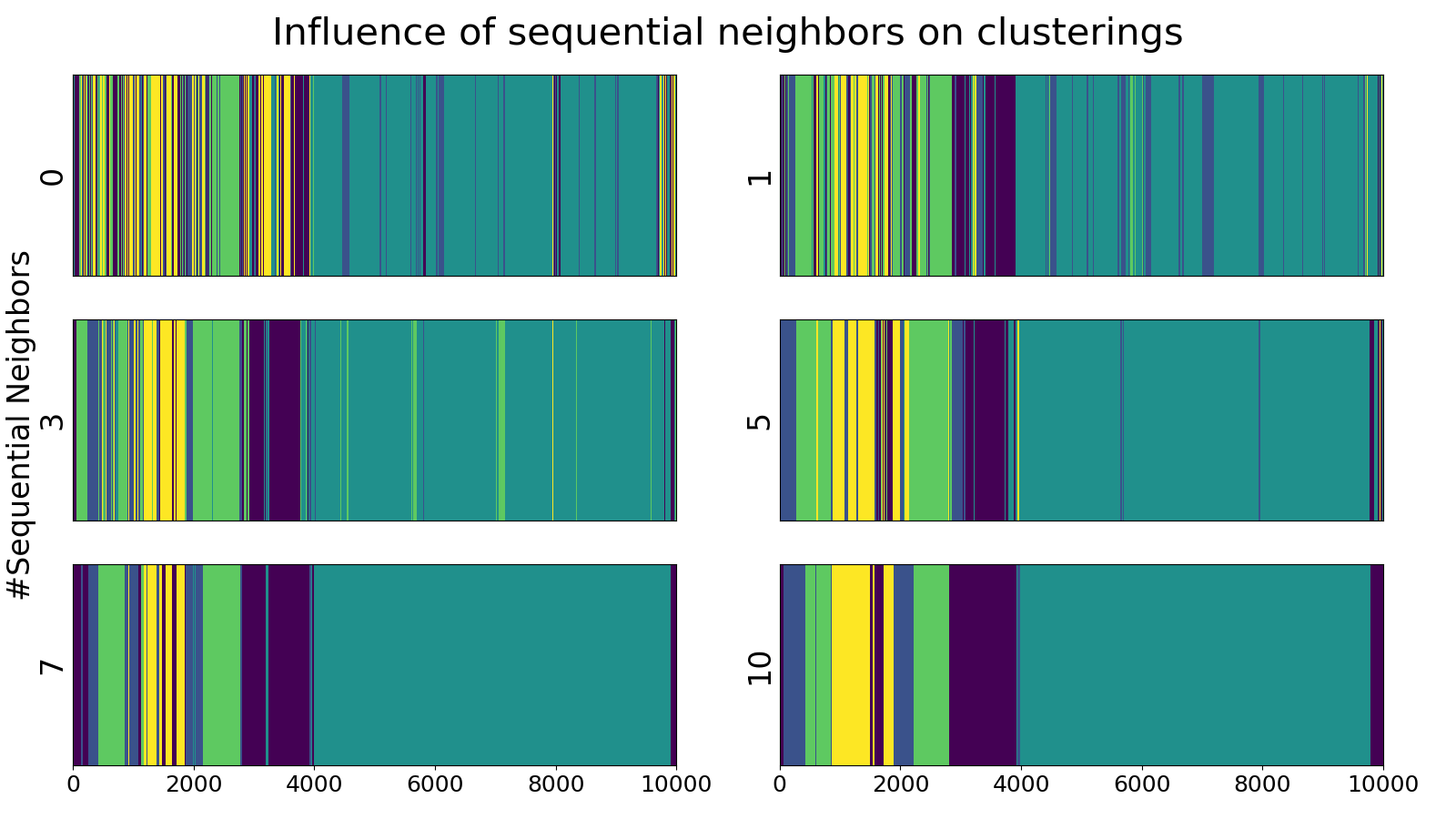}
    \caption{Varying number of sequential neighbors for 2F4K protein  clustered into 5 clusters. Time steps are represented along the x-axis, clusters are implied by color.}
    \label{fig:tsc_param_seq_neighbors_clustering}
\end{figure}

Setting the number of sequential neighbors $s$ in the temporal regularization to 0, the temporal aspect is not considered, with growing $s$ longer temporal relationships are captured. The scores for $s\in\{0,...,12\}$ are shown in Figure \ref{fig:tsc_param_seq_neighbors} for different MSM lags (columns) and number of clusters (rows). 
The number of sequential neighbors $s$ clearly affects clustering performance, with too few or too neighbors reducing performance, but with a broad stable range where temporal relationships are successfully captured.
Figure \ref{fig:tsc_param_seq_neighbors_clustering} shows the clustering of a 2F4K protein trajectory with a long folded state into 5 clusters represented by different colors. With no (top) or few (center) sequential neighbors, time intervals comprised by a cluster can be very short.
Increasing the number of sequential neighbors (bottom) leads to continuous clusters with mostly uninterrupted blocks.  

\subsubsection{$\lambda$ parameters} \label{exp:lambda}
\begin{figure}[ht]
    \centering
    \includegraphics[width=\linewidth]{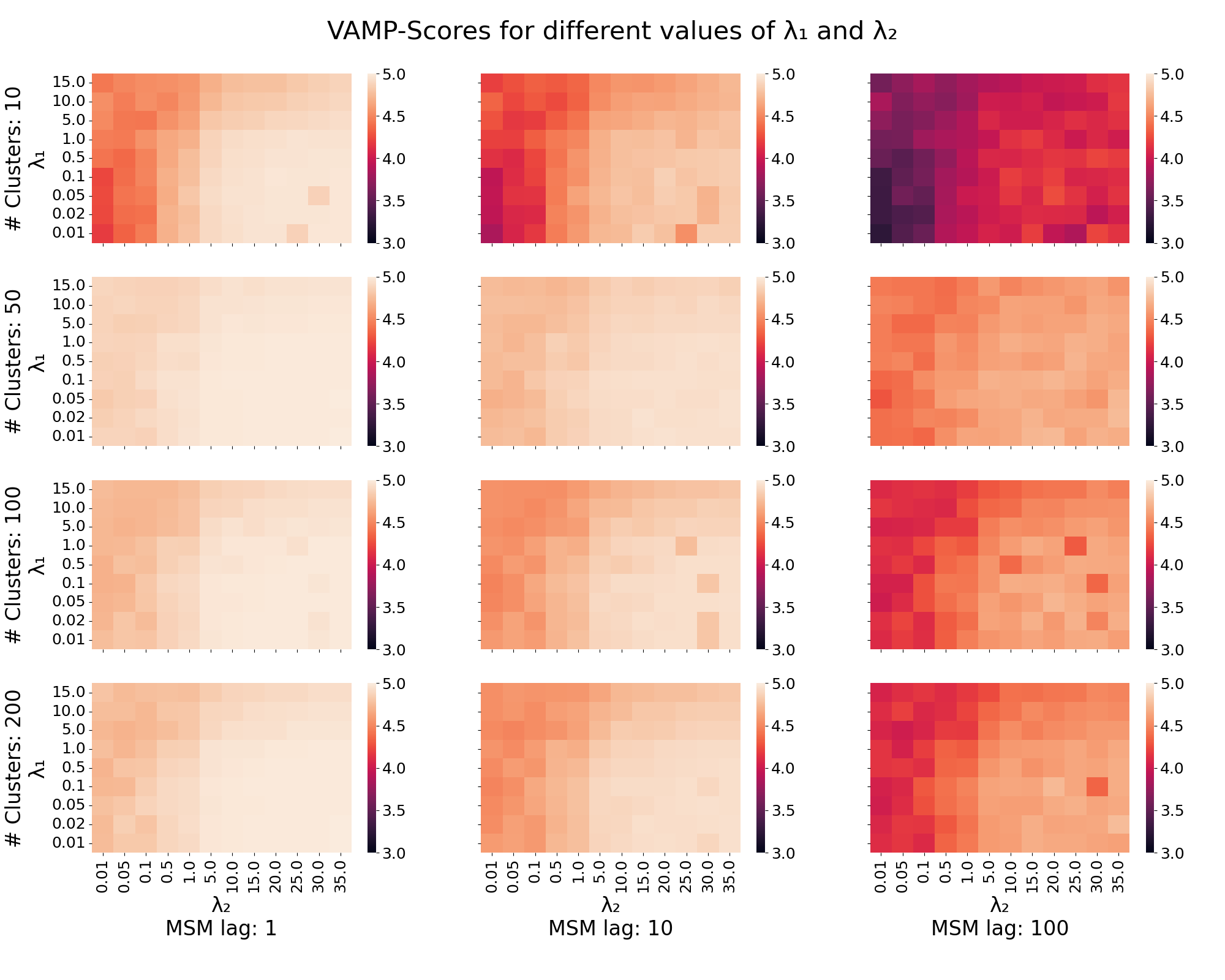}
    \caption{Heatmap of VAMP-scores (the higher/lighter, the better; 
    scale from 3 to 5 for visibility); best results for values around 0.01 for $\lambda_1$ and around 15 for $\lambda_2$.}
    \label{fig:tsc_lambda}
\end{figure}

We study settings of the tradeoff parameters  $\lambda_1 \in \{0.01, 0.05, \dotsc, 15\}$ for sparsity  of the coding matrix $Z$, $\lambda_2 \in \{0.01, 0.05, \dotsc, 35\}$ for the temporal regularization function.
Figure \ref{fig:tsc_lambda} shows that best performance is achieved for low values of $\lambda_1$  and higher values of $\lambda_2$ (light areas in the heatmap). This shows that temporal regularization is the more important factor for optimal clustering performance.

\subsubsection{Learning rates} \label{exp:alpha_beta}
\begin{figure}
    \centering
    \includegraphics[width=\linewidth, trim= {0 20 20 20},clip]{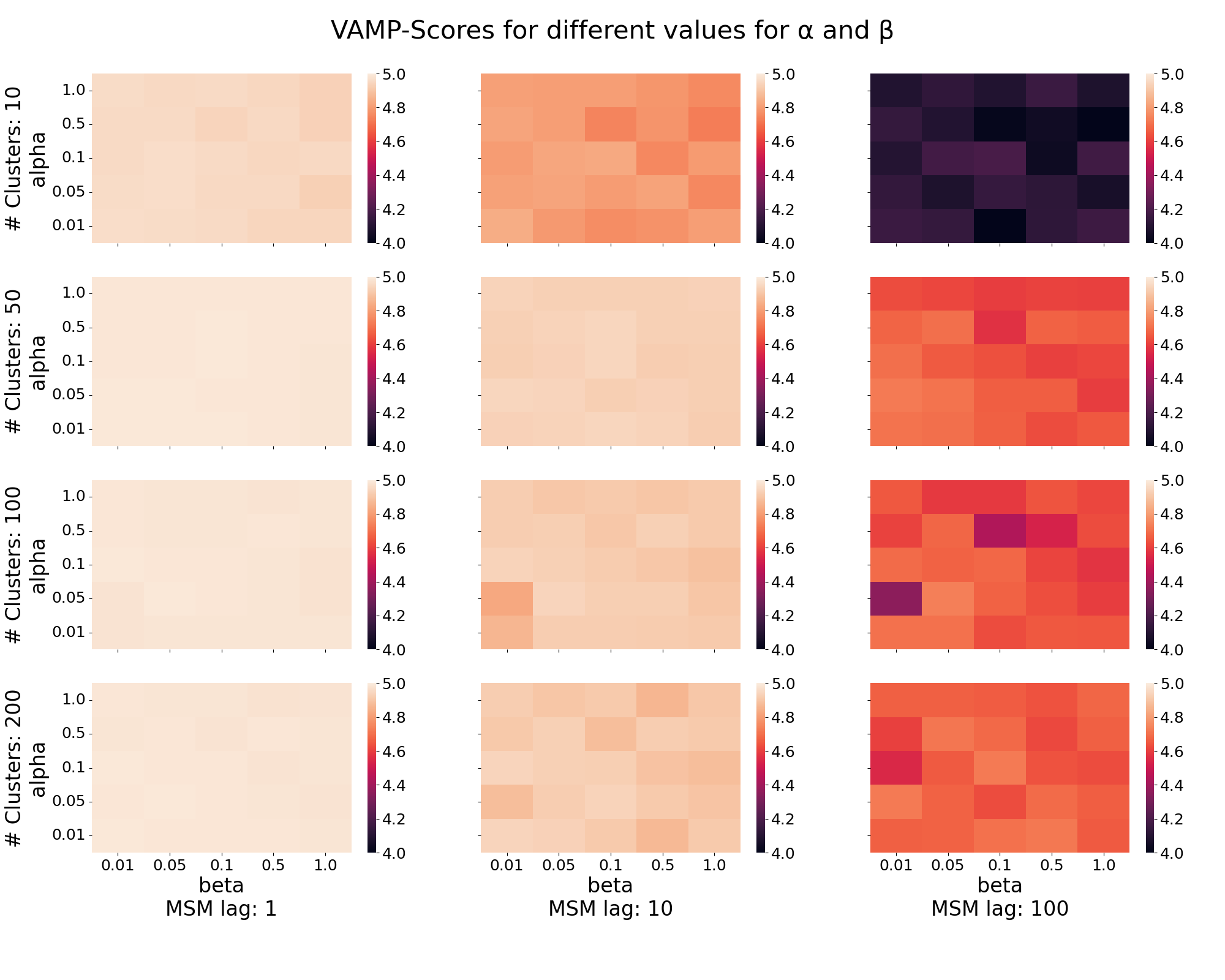}
    \caption{VAMP-scores for varying $\alpha$ and $\beta$: almost no impact on clustering performance.}
    \label{fig:tsc_param_alpha_beta}
\end{figure}

$\alpha, \beta \in\{0.01, 0.05, \dotsc, 15\}$ 
define how much Lagrangian multipliers $\Pi$ and $\Lambda$ change in each iteration of the algorithm.
Heat maps in Figure \ref{fig:tsc_param_alpha_beta} show that the results are stable across parameter settings, with VAMP2-scores in each heatmap relatively close, meaning our default values work well.

\subsubsection{Weighting mode} \label{exp:weight}
\begin{figure}
    \centering
    \includegraphics[width=\linewidth, trim= {0 100 20 20},clip]{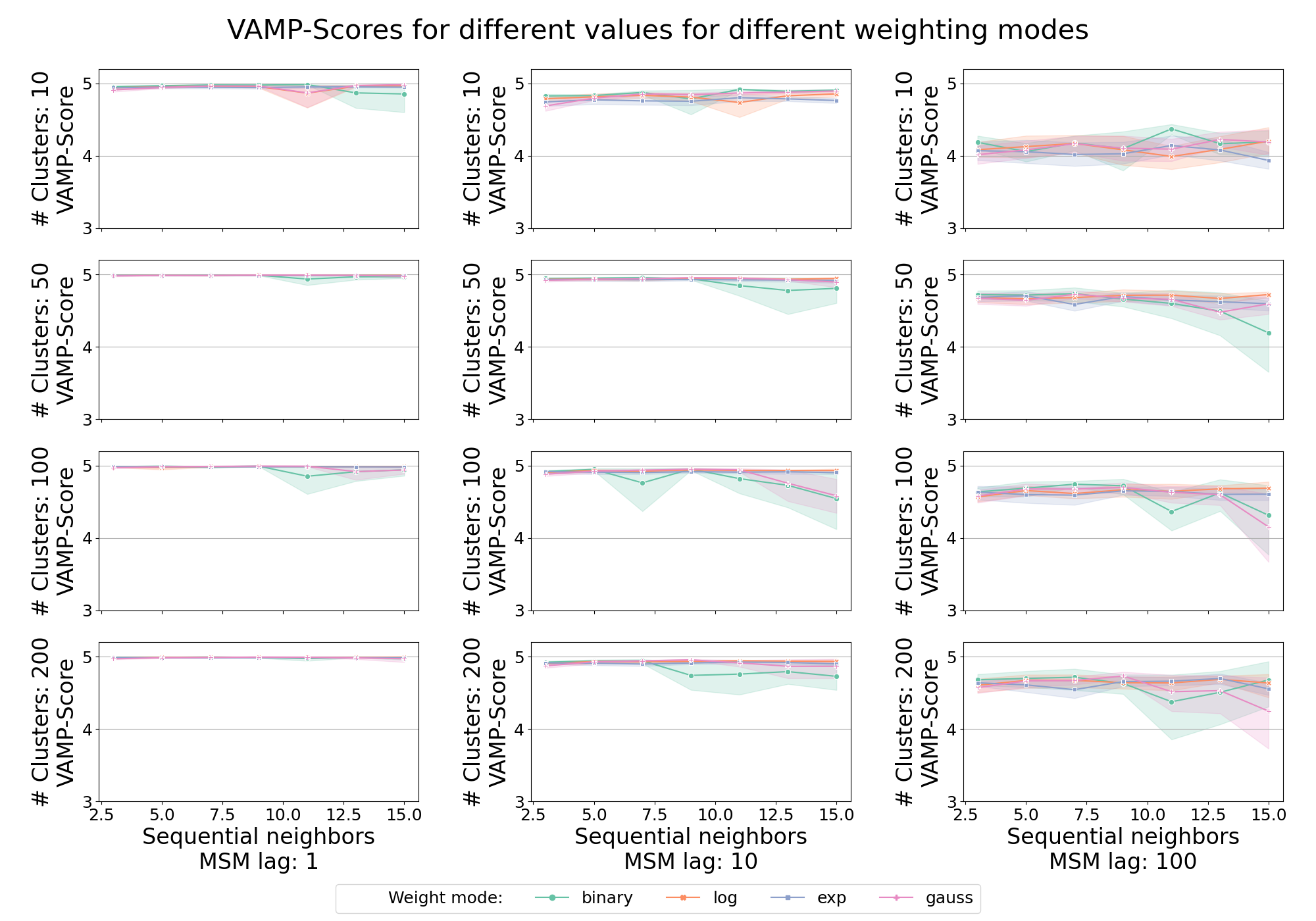}
    \includegraphics[width=\linewidth, trim= {360 0 360 960},clip]{images/experiments/TSC-parameters/plot_weight_mode.png}
    \caption{VAMP-scores for  weighting modes of temporal Laplacian}
    \label{fig:tsc_param_weight}
\end{figure}

All weighting modes for the Laplacian regularization function: binary, Gaussian, logarithmic, and exponential, are tested for different numbers of sequential neighbors.
In Figure \ref{fig:tsc_param_weight} we see that for few sequential neighbors, the performance for all weighting modes is similar. For many sequential neighbors, performance starts to decrease. In most cases, the performance when using the binary weighting mode decreases earlier than for the other methods, but all reliably reach the same maximal score.

\subsubsection{Feature comparison} \label{sec:feature_comp}
\begin{figure*}
    \centering
    \includegraphics[width=.7\textwidth, trim={70 130 70 225}, clip]{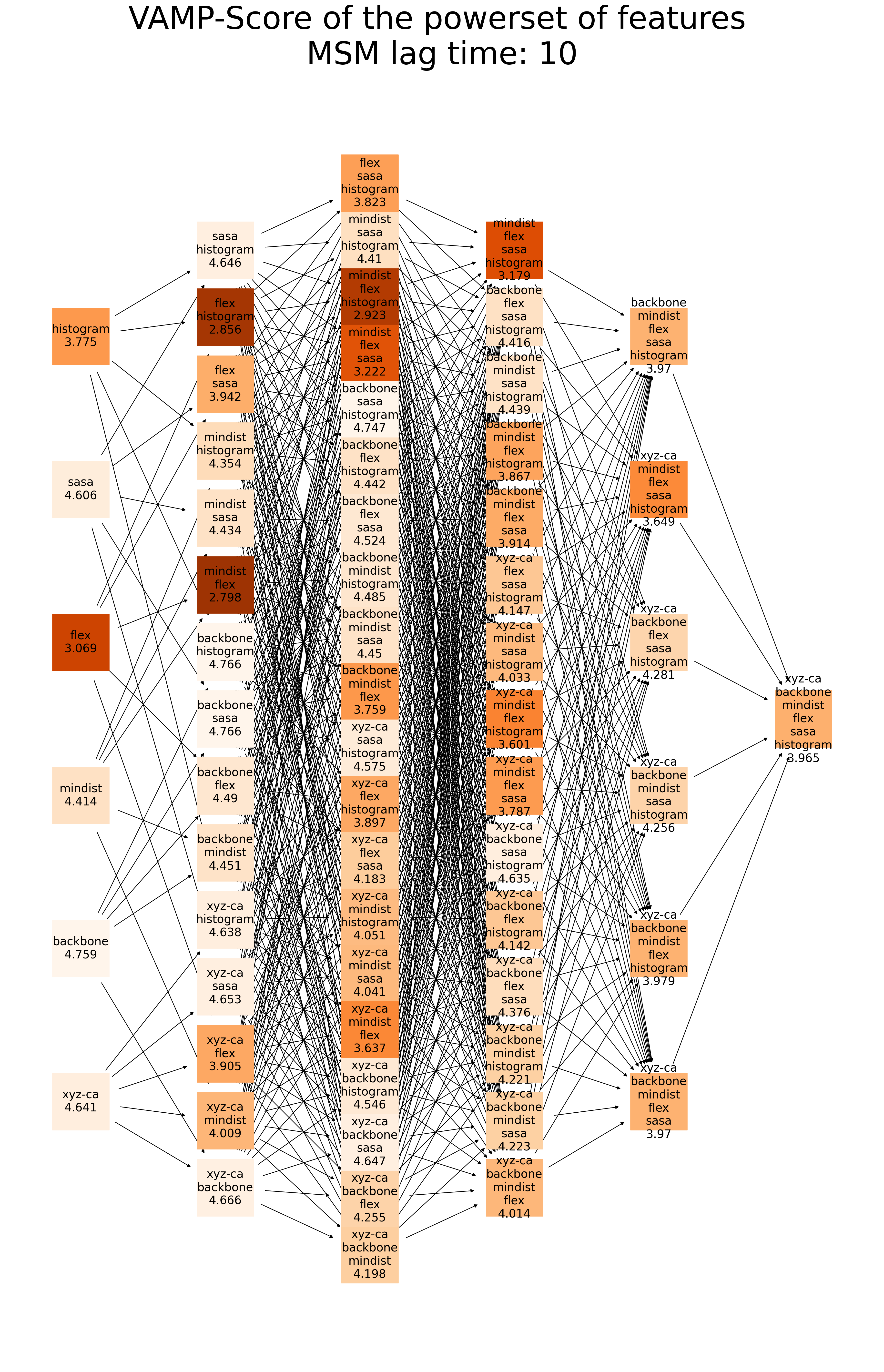}
    \caption{VAMP2-Scores for different featurizations (average over 10 trajectories for powerset of 6 features). Number of applied features increases from left to right. Colors correspond to the VAMP2-Scores, lighter values are better. }
    \label{fig:powerset}
\end{figure*}

We analyze the above described features and their combinations via simple concatenation. 
Figure \ref{fig:powerset} 
shows the average VAMP2-score for 10 trajectories of the 2WAX-protein, using the five largest eigenvalues for scoring. 

\begin{figure}
    \centering
    \includegraphics[width=.6\columnwidth]{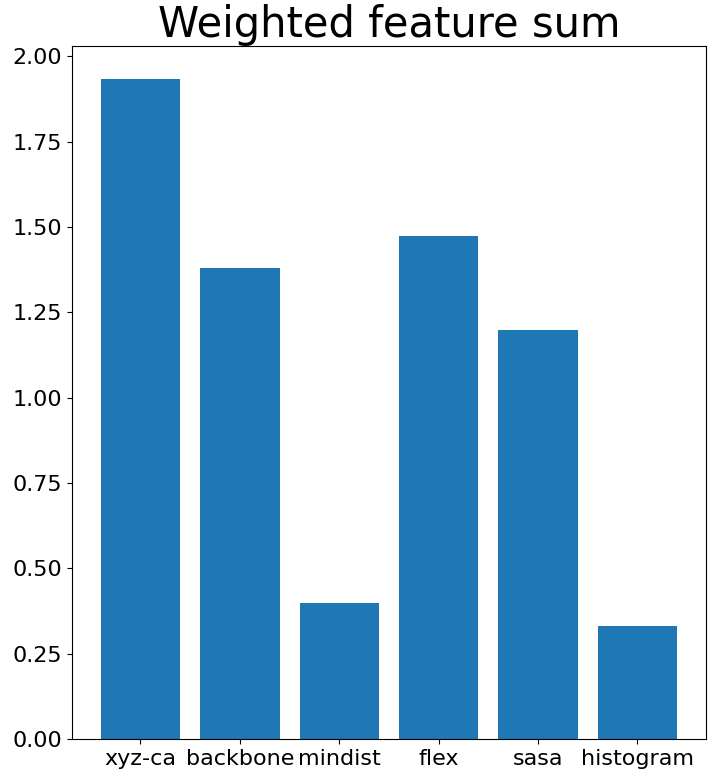}
    \caption{Weights (=relevance) of each feature according to the dictionary learned by MOSCITO when given the full set of features. }\label{fig:feature_weights}
\end{figure}

Best results for individual features are achieved with backbone torsions. Overall, combinations of backbone torsions, SASA, and shape histograms yield the best results. Figure \ref{fig:feature_weights} shows the importance that MOSCITO gives to each of the features in its dictionary when clustering the full-dimensional feature space. While, e.g., the feature "flex" (flexible torsions) yields the worst VAMP2 score (left), it still has a high influence on the results (right), explaining the suboptimal VAMP2 score when using the full feature set.

\subsection{Clustering performance} \label{sec:exp_comp_all}
In this section, the clustering performance of MOSCITO is compared to that of state-of-the-art approaches that are currently used in the field. We compare to PCA with $k$-Means and to TICA with $k$-Means because of their widespread use~\cite{pyemma_tutorial}. Additionally, we compare against the sparse spectral clustering algorithm \cite{SSC_MD}, a subspace clustering approach for molecular dynamics data.
For PCA and TICA, the default parameter settings of PyEmma \cite{pyemma_tutorial} were used. SSC is a subspace clustering algorithm, thus it does not need any dimensionality reduction in the preprocessing. 

A general challenge in this kind of evaluation is the fact that no ground truth is given. As such, clustering performance is compared by constructing a MSM and scoring it using the VAMP2-score, in order to obtain quantitative insights. Still, visual inspection plays an important role in assessing the clustering performance, in particular, the temporal connectivity of clusters.

\begin{figure}
    \centering
    \includegraphics[width=0.99\columnwidth, trim={25 100 0 0}, clip]{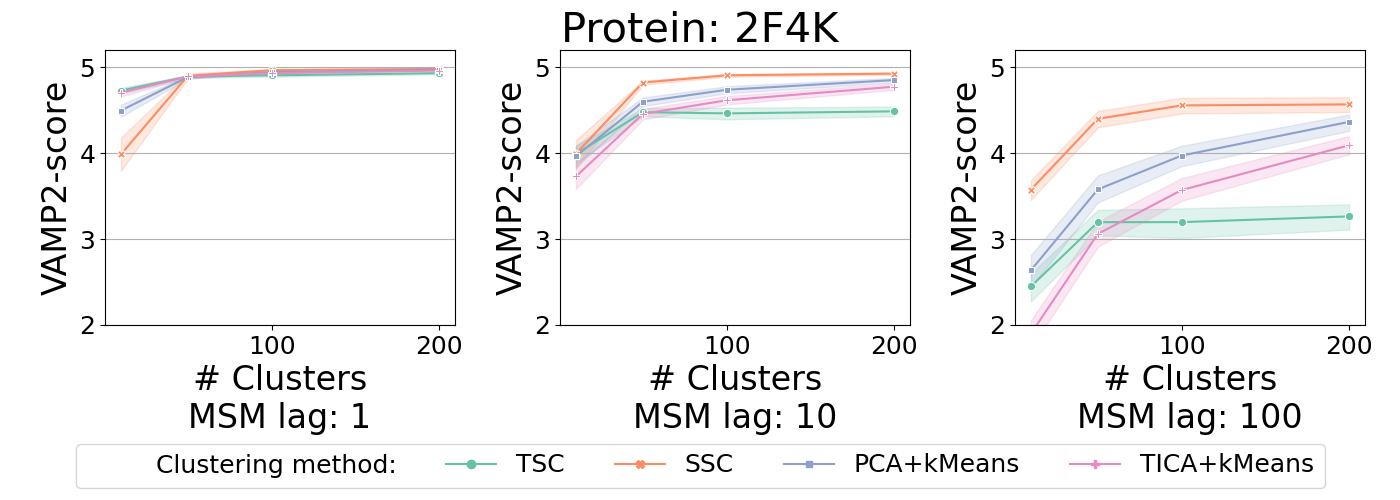}
    \centering
    \includegraphics[width=0.99\columnwidth, trim={0 100 0 0}, clip]{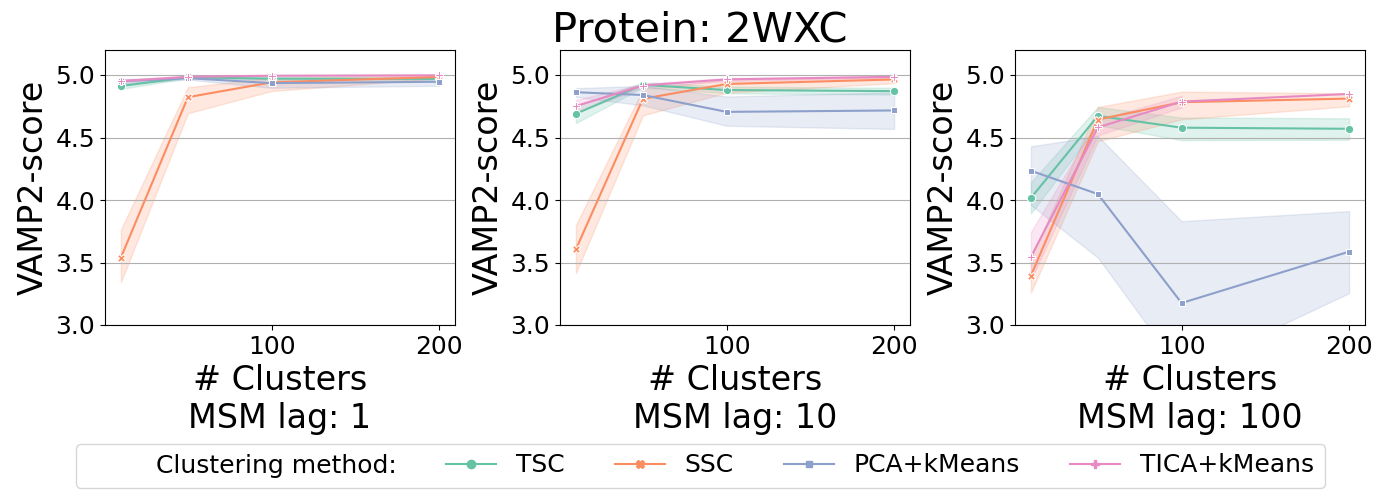}
    \centering
    \includegraphics[width=0.99\columnwidth, trim={0 100 0 0}, clip]{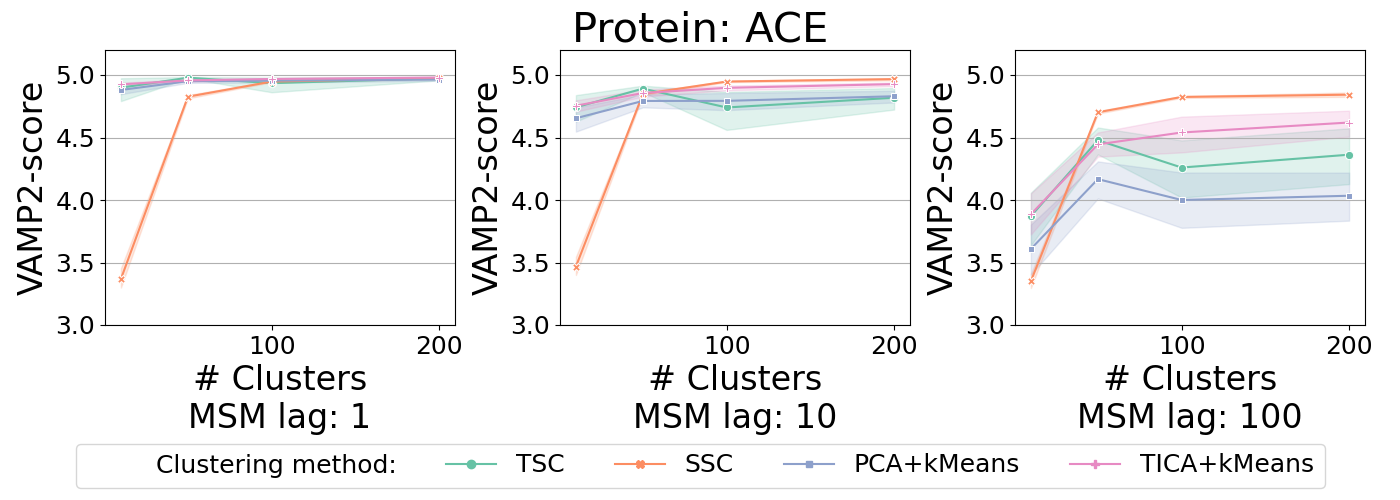}
    \includegraphics[width=0.99\columnwidth, trim={0 100 0 0}, clip]{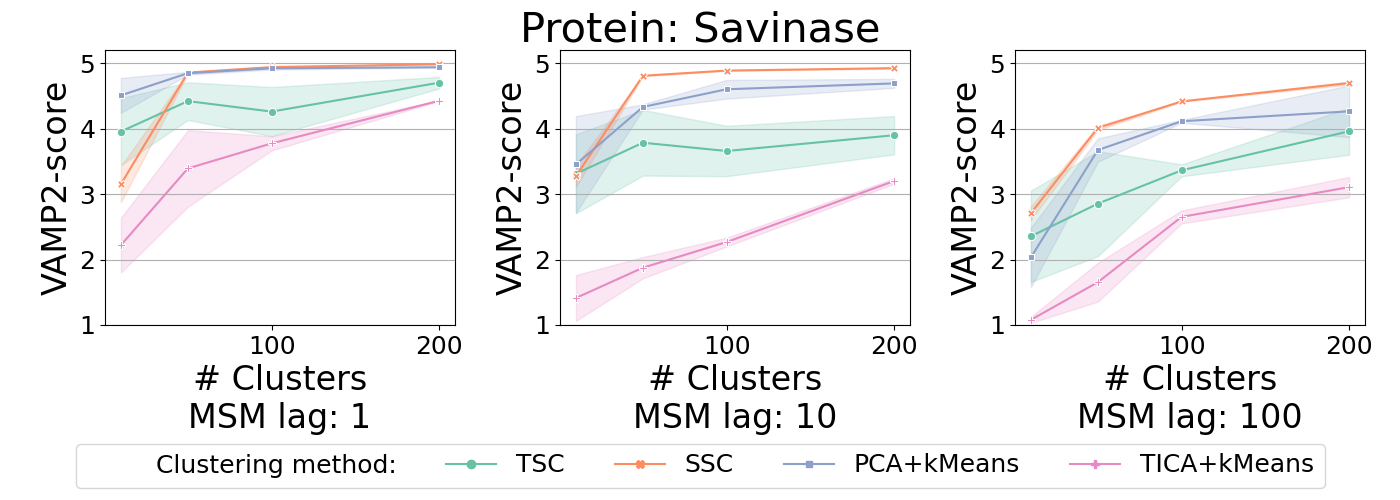}
    \includegraphics[width=0.99\columnwidth, trim={0 0 0 0}, clip]{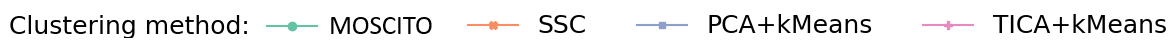}
    \caption{Mean VAMP2-Scores for 2F4K, 2WXC, ACE, and Savinase (top to bottom)}
    \label{fig:scores_all_proteins}
\end{figure}

\paragraph*{Quantitative analysis}
Figure \ref{fig:scores_all_proteins} shows the average VAMP2-score for 20 trajectories of the 2F4K protein in the first row. For MSM lag time 1, the performance of all methods is comparable with the exception of 10 clusters. There, MOSCITO and TICA + $k$-Means deliver the best performances. For longer MSM lag times, the scores separate with SSC performing better.

In the second row of Figure \ref{fig:scores_all_proteins}, we see the average VAMP2-score for 10 trajectories of the 2WXC protein. For the MSM lag times 1 and 10, the performance for all methods is relatively similar for large cluster sizes. For small cluster sizes, SSC performs poorly, while the other methods already reach close to their maximal score for only 10 clusters. For MSM lag time of 100, PCA + $k$-Means drastically drops in performance, while the other methods maintain comparable scores to each other.

The third row of Figure \ref{fig:scores_all_proteins} shows the average VAMP2-score for 28 trajectories of the Ace protein. Similar to the results of 2WXC, the performance for a MSM lag time of 1 and 10 is comparable between all methods for a large number of clusters. For few clusters, SSC again performs significantly worse. For MSM lag time of 100, the gap between the methods grows, with SSC performing best.

The last row of Figure \ref{fig:scores_all_proteins} shows the average VAMP2-score for 2 trajectories of Savinase. For all MSM lag times, SSC performs best while TICA + $k$-Means performs worst. Unlike for the other proteins, all methods require a large number of clusters to reach the maximal VAMP2-score. This might be due to the large size of Savinase and shorter trajectories compared to the other proteins.

\begin{figure}
    \centering
    \includegraphics[width=1\columnwidth]{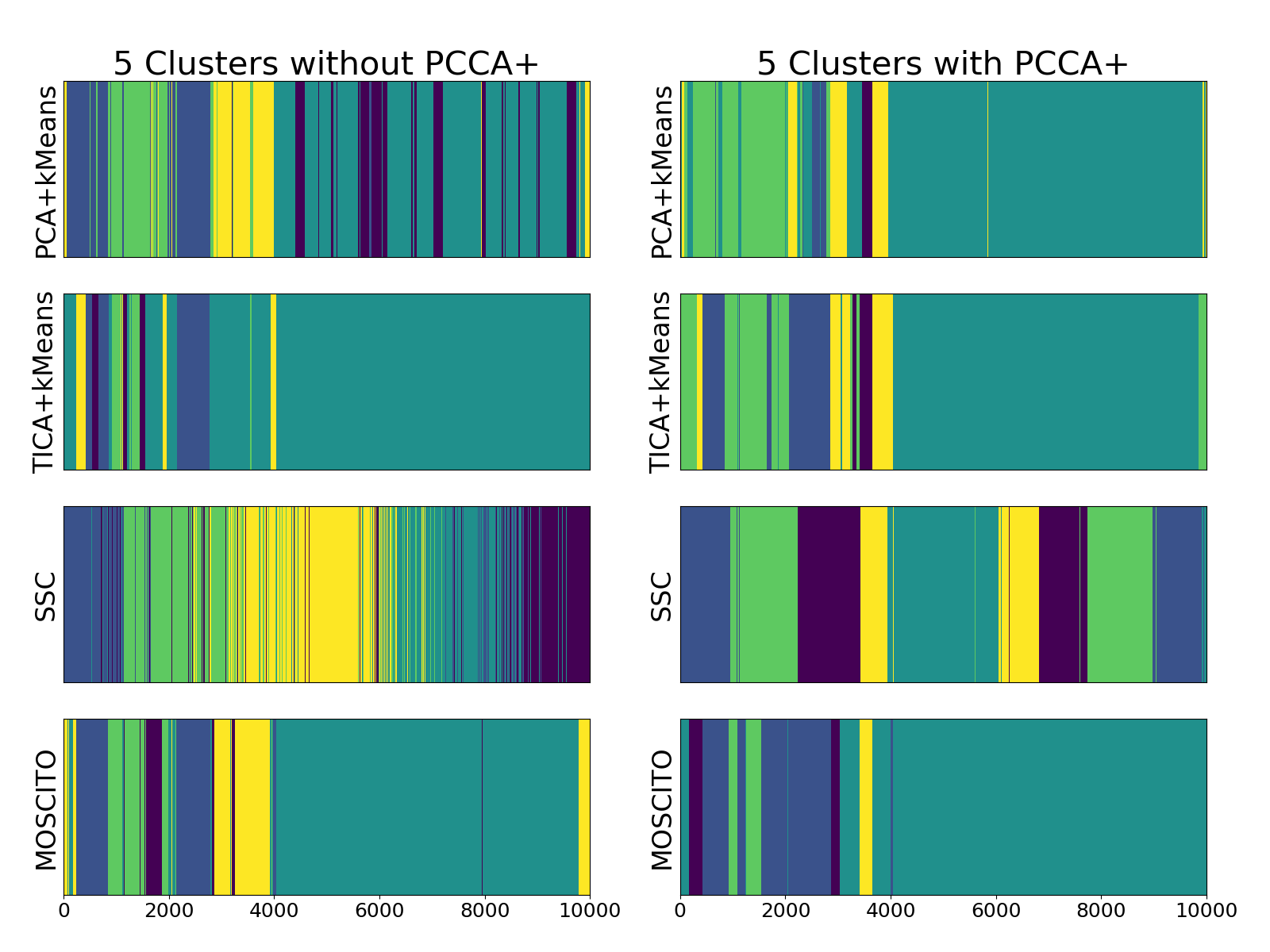}
    \caption{Clustering impact of PCCA+ on a 2F4K trajectory (each color represents a cluster; left: clustering the trajectory directly into 5 clusters; right: coarse-graining 50 clusters into 5 clusters using PCCA+).}
    \label{fig:comp_pcca}
\end{figure}

\paragraph*{Qualitative analysis}
The purpose of modeling temporal aspects is to obtain clusters with larger continuous segments, ideally retrieving macrostates directly.  Figure \ref{fig:comp_pcca} visually compares the segmentation of clustering methods, either by directly clustering the trajectory into five clusters or, alternatively by coarse-graining 50 clusters into five clusters using PCCA+~\cite{pcca} (see \Cref{chap:rel_work}). We regard 
a 2F4K trajectory with a known long folded segment.

Directly clustering, i.e. without PCCA+, PCA + $k$-Means and SSC struggle to find the large folded state as one large block. MOSCITO and TICA+$k$-Means find similar clusterings and succeed to uncover the large folded state. When applying PCCA+, PCA + $k$-Means, TICA + $k$-Means, and MOSCITO generate similar clusterings, comparable to the ones of MOSCITO and TICA + $k$-Means without PCCA+. In this specific case, SSC fails to find the large segment, which is reflected in the respective VAMP2-score for this specific trajectory which is unusually low.

\subsection{Clustering runtime} \label{exp:runtime}
\begin{figure}[bt]
    \centering
    \begin{minipage}{0.48\columnwidth}
        \centering
        \includegraphics[width=\linewidth]{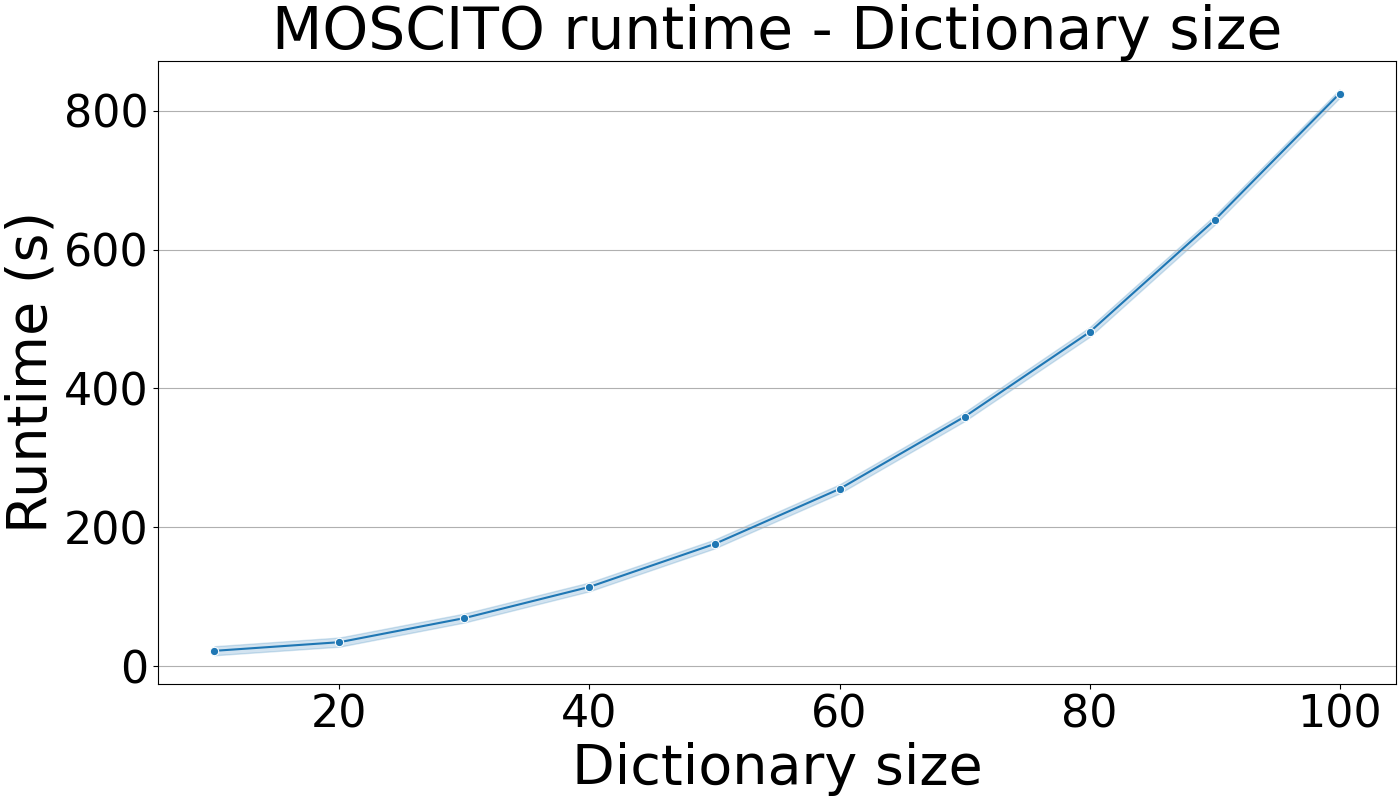}
        \label{fig:runtime_moscito_d}
    \end{minipage}
    \hfill
    \begin{minipage}{0.48\columnwidth}
        \centering
        \includegraphics[width=\linewidth]{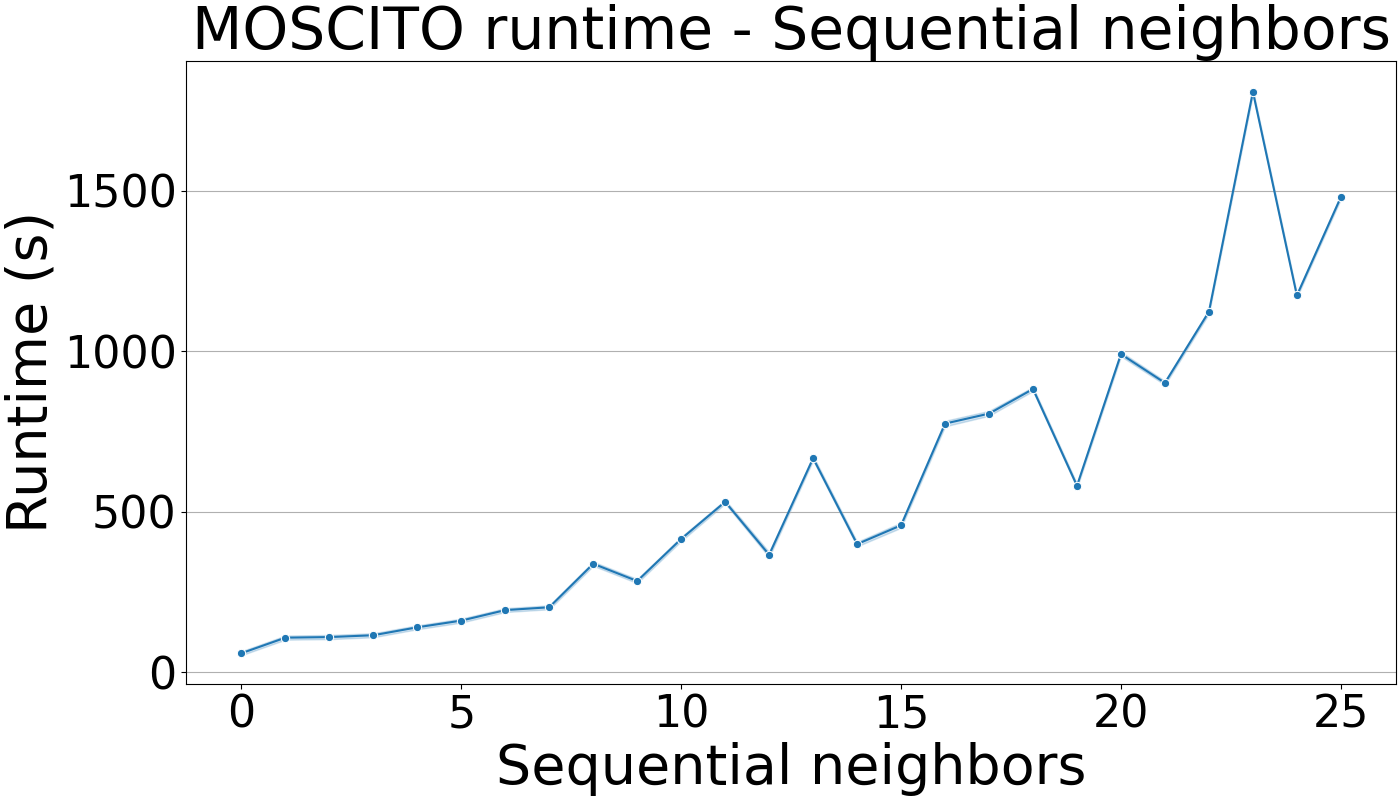}
        \label{fig:runtime_moscito_s}
    \end{minipage}
    \caption{MOSCITO runtime (left: dictionary sizes; right: number of sequential neighbors).}
    \label{fig:runtime_analysis}
\end{figure}

In this section, the runtime of MOSCITO is studied and compared to the runtime of PCA + $k$-Means, TICA + $k$-Means, and SSC. The runtimes for MOSCITO are measured for 20 iterations. The following runtimes are the average of 10 2WXC-trajectories clustered into ten clusters, using all atom coordinates as features. 
The main factors influencing the runtime of MOSCITO, besides the size of the input data, are the dictionary size $d$ and the number of sequential neighbors $s$. The effect of different dictionary sizes $[10, ..., 100]$ on the runtime is shown left in Figure \ref{fig:runtime_analysis}. The effect of different numbers of sequential neighbors $[0, ..., 25]$ is on the right. 

The runtime of MOSCITO grows with increasing dictionary sizes. For optimal cluster performance, a large enough dictionary size has to be selected. Large values do not affect clustering performance negatively, but directly impact runtime. A dictionary size between 60 and 80 provides a good tradeoff.
The runtime generally grows with increasing number of regarded sequential neighbors. 
The overall run-to-run variance for the different trajectories is close to zero, indicating a stable runtime for a fixed number of iterations. 

\begin{figure}
   \centering   \includegraphics[width=.7\columnwidth]{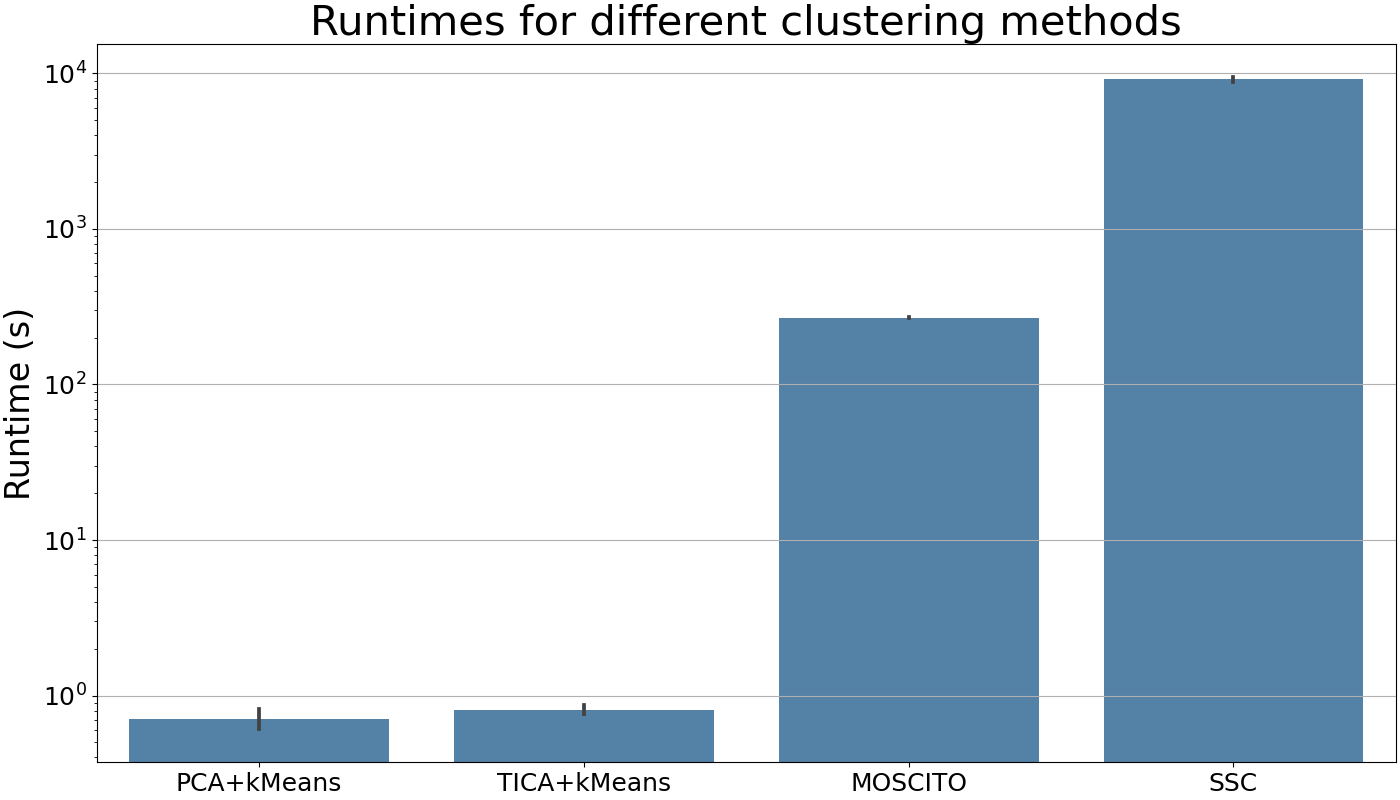}
   \caption{Runtime for PCA+k-Means, TICA+k-Means, MOSTCITO, SSC (logarithmic scale on y-axis for readability)}
   \label{fig:runtime_all}
\end{figure}

For the comparison of the different clustering methods, backbone torsions were used. The comparison between the different clustering algorithms is shown in Figure \ref{fig:runtime_all}.
PCA + $k$-Means and TICA + $k$-Means are both much faster than the subspace clustering methods.  Compared to SSC, MOSCITO is substantially faster. 
\subsection{Discussion} \label{sec:exp_discussion}
We conclude that MOSCITO provides at least comparable performance to state-of-the-art clustering methods for MD data. We evaluated the clustering performance by the clusters' suitability as Markov states using the VAMP2-score. The visual analysis of MOSCITOs clustering results shows a better clustering for higher numbers of sequential neighbors. Thus, MOSCITO is a very promising approach for creating MSMs in a one-step-approach.

MOSCITO's parameter robustness shows that there are good tradeoffs in dictionary size between runtime and clustering performance. The number of sequential neighbors matters, and our experiments indicate a good default value is between 3 and 5 to capture nearby temporal relations. 
In terms of feature extraction, MOSCITO on backbone torsions leads to best results, followed by C$\alpha$ coordinates and SASA. Combining features does not improve the overall clustering performance. 
While simpler non-subspace methods are even faster, MOSCITO provides a faster runtime than SSC, and state-of-the-art quantitative results, with qualitative analysis showing the promise of the approach.

\section{Related Work} \label{chap:rel_work}
Distance-based clustering algorithms suffer from the curse of dimensionality when it comes to high-dimensional data like MD data.
Thus, the dimensionality is often reduced in a preprocessing step, e.g., with principal component analysis (PCA) or time-lagged independent component analysis (TICA) \cite{molgedey1994separation}. Both are widely used for analyzing MD data \cite{pca_usage, tica_usage}. 
PCA transforms the data onto the axes with the highest variance, also known as principal components. 
PCA is a commonly used for MD data, e.g., in PyEMMA \cite{scherer_pyemma_2015}, MSMBuilder \footnote{\url{http://www.msmbuilder.org}}, or MDAnalysis \footnote{\url{https://www.mdanalysis.org/}}.
TICA (Time-lagged independent component analysis) \cite{molgedey1994separation} linearly projects the data to independent, uncorrelated components (IC). 
The ICs are computed by maximizing the autocovariance for a fixed lagtime $\tau$, and are frequently used for MD data (\cite{schwantes2013improvements}, \cite{perez2013identification}).

State-of-the-art methods often follow a two-step approach, where the trajectories are discretized into a large number of \textit{microstates}, and subsequently combined to the relevant \textit{macrostates}.
Since the transitions between the metastable states are taking place over multiple timesteps
it is hard to assign time steps that are part of a transition to a metastable state clearly. 
Perron-Cluster Cluster Analysis (PCCA)~(\cite{pccaBasis},\cite{pcca}) and a more robust version PCCA+~(\cite{weber2002characterization}, \cite{deuflhard2005robust}) fuzzily assign the microstates to their corresponding macrostates based on spectral methods for transition matrices:

The transition matrix $P$ of a decomposable MSM with $n_c$ macrostates can be reordered to a block-diagonal matrix $\tilde P$ with $n_c$ blocks. 
For the matrix $X$ containing the $n_c$ dominant eigenvectors of $\tilde P$, the rows belonging to states in the same block are identical \cite{Frank2022}. 
These rows represent the $n_c$ corners of a ($n_c$-1) dimensional simplex.

For MSMs with transition states, i.e., nearly decomposable MSMs, the structure of the simplex is only slightly perturbed. To assign the microstates to the macrostates, they are fuzzily assigned to the $n_c$ corners of the simplex based on their position in the simplex. The goal is to find a linear transformation $A\in \mathbb{R}^{n_c\times n_c}$ so that $\chi=XA$ where $\chi_i$ are the membership vectors. Those memberships all have to be positive and sum up to 1. 
The resulting macrostates generally cannot be considered to be a Markov chain \cite{pcca}.

Time series data can be found in many fields, like traffic analysis, geology, computer vision, or molecular dynamics data. Many existing subspace clustering algorithms do not take the sequential properties of time series data into account \cite{wu2015ordered}. One of the first subspace clustering methods taking advantage of sequential relationships in data was \textit{SpatSC} introduced by \cite{guo2014spatial}. A penalty term used to preserve sequential relationships in the learned affinity graph was added to the subspace clustering problem. Based on the same idea, \textit{OSC}, a more robust subspace clustering method for sequential data was introduced by \cite{tierney2014subspace}. 
\textit{TSC} \cite{li2015tsc} further improved on the idea by defining a Laplacian regularization to encode the temporal information. 
Our method MOSCITO builds upon \textit{TSC}. 

To the best of our knowledge, only one subspace clustering method has been used on molecular dynamics data, namely SSC~\cite{SSC_MD}, which was first introduced in the field of image processing \cite{SSC}. We include SSC in our experiments as comparative method, thus we introduce it in detail in the following. 
Note that SSC does not take advantage of any properties that distinguish time series from other high-dimensional data, leading to suboptimal use of the available information in MD data.
The main idea behind SSC is to take advantage of the self-expressiveness of the data, i.e., assuming that every point $y_i\in Y$ can be rewritten as a linear combination of other points in the dataset. Each datapoint $y\in \cup_{i=1}^nS_i$ can be written as
\begin{equation}
    y_i=Yc_i,\;\; c_{ii}=0,
    \label{eq:ssc}
\end{equation}
where $Y=[y_1,...,y_N]$ and $c_i:=[c_{i1}\; c_{i2}\; ...\; c_{iN}]^T$ with the constraint $c_{ii}=0$ eliminating the trivial solution of representing the point as itself. 
Since each subspace usually has more data points than dimensions the self-expressive dictionary $Y$ is generally not unique. 
Among all those possible solutions for Equation \ref{eq:ssc} there is a sparse solution so that the non-zero entries of $c_i$ correspond to datapoints in the same subspace as $y_i$. This solution can be found by minimizing the $l_1$-norm of $c_i$, which can be solved efficiently and is known to prefer sparse solutions \cite{SSC}.

These sparse representations are transferred to an undirected graph where points that are part of another point's sparse representation are connected. Based on this graph, spectral clustering detects the desired subspace clusters.

\section{Conclusion} \label{chap:conclusion}
We introduced MOSCITO, a subspace clustering method for high-dimensional molecular dynamics data. It uses temporal regularization to capture the relationship of sequential neighbors in time-series data.
While traditional and state-of-the-art methods rely on coarse-graining using PCCA+ in a two-step approach, MOSCITO achieves similar overall performance in a single clustering step.
Especially for small numbers of clusters, the temporal regularization in MOSCITO yields clusters consisting of connected time steps, fitting the real world more closely than alternative methods without relying on any postprocessing steps. 
Compared to current subspace clustering methods on MD data, MOSCITO provides a better runtime.
In future work, the integration of a multi-view approach into MOSCITO might take advantage of the additional information provided by using multiple features.

\bibliographystyle{ACM-Reference-Format}
\bibliography{references}

\end{document}